\documentclass[conference,compsoc]{IEEEtran}
\IEEEoverridecommandlockouts

\ifCLASSOPTIONcompsoc
  \usepackage[nocompress]{cite}
\else
  \usepackage{cite}
\fi
\ifCLASSINFOpdf
   \usepackage[pdftex]{graphicx}
\else
\fi
\usepackage{amsmath}
\usepackage{color}
\usepackage{array}

\ifCLASSOPTIONcompsoc
  \usepackage[caption=false,font=footnotesize,labelfont=sf,textfont=sf]{subfig}
\else
  \usepackage[caption=false,font=footnotesize]{subfig}
\fi

\usepackage{stfloats}
\usepackage{float}
\usepackage{xcolor}

\usepackage[hyphens]{url}

\usepackage{color,soul}

\begin{document}
%
\title{Hand Gesture Controlled Drones: An Open Source Library}

\author{\IEEEauthorblockN{Kathiravan Natarajan\IEEEauthorrefmark{2}}
\IEEEauthorblockA{Student IEEE Member\\
Department of Computer Science\\
Texas A\&M University -- Commerce \\
Commerce, Texas 75429\\
sunkathirav@gmail.com
}
\thanks{\dag Supported by Texas A\&M University--Commerce Graduate School and Department of Computer Science}
\and 
\IEEEauthorblockN{Truong-Huy D. Nguyen\IEEEauthorrefmark{1}}
\IEEEauthorblockA{IEEE Member\\Department of Computer \\
and Information Sciences\\
Fordham University \\
Bronx, New York 10458 \\
tnguyen88@fordham.edu
}
\thanks{$*$Corresponding author}
\and
\IEEEauthorblockN{Mutlu Mete}
\IEEEauthorblockA{IEEE Member\\ 
Department of Computer Science\\
Texas A\&M University -- Commerce \\
Commerce, Texas 75429 \\
mutlu.mete@tamuc.edu
}

}


%


\maketitle

\begin{abstract}
Drones are conventionally controlled using joysticks, remote controllers, mobile applications, and embedded computers. A few significant issues with these approaches are that drone control is limited by the range of electromagnetic radiation and susceptible to interference noise. In this study we propose the use of hand gestures as a method to control drones. We investigate the use of computer vision methods to develop an intuitive way of agent-less communication between a drone and its operator. Computer vision-based methods rely on the ability of a drone’s camera to capture surrounding images and use pattern recognition to translate images to meaningful and/or actionable information. The proposed framework involves a few key parts toward an ultimate action to be taken. They are: image segregation from the video streams of front camera, creating a robust and reliable image recognition based on segregated images, and finally conversion of classified gestures into actionable drone movement, such as takeoff, landing, hovering and so forth. A set of five gestures are studied in this work. Haar feature-based AdaBoost classifier\cite{viola2001rapid} is employed for gesture recognition. We also envisage safety of the operator and drone's action calculating the distance based on computer vision for this task.  A series of experiments are conducted to measure gesture recognition accuracies considering the major scene variabilities, illumination, background, and distance. Classification accuracies show that well-lit, clear background, and within 3 ft gestures are recognized correctly over 90\%. Limitations of current framework and feasible solutions for better gesture recognition are discussed, too.  The software library we developed,and hand gesture datasets are open-sourced at project website.

\end{abstract}


%
\IEEEpeerreviewmaketitle

\section{Introduction}
Drones, also known as unmanned aerial vehicles, are on the rise in recreational and in a wide range of industrial applications, such as security, defense, agriculture, energy, insurance and hydrology. Drones are essentially special flying robots that perform functionalities like capturing images, recording videos and sensing multimodal data from its environment. There are two types of drones based on their shape and size, fixed-wing and multirotor. Because of their versatility and small size, multirotor drones can operate where humans cannot, collect multimodal data, and intervene in occasions. Moreover, with the use of a guard hull, multirotor drones are very sturdy in collisions, which make them even more valuable for exploring uncharted areas. At present, flying robots are used in different businesses like parcel delivery systems \cite{gatteschi2015}. For example, companies like Amazon Prime and UPS are using multirotor drones to deliver their parcels. New York Police Department uses quadcopters in crime prevention \cite{NYPD}. For the purposes of agriculture monitoring, for instance, the use of multiple sensors such as video and thermal infrared cameras are of benefit \cite{turner2014spatial}. Drones are especially useful in risky missions. For the sake of clarity in the rest of this work, we define \emph{a drone} as a multirotor flying robot, excluding fixed-wings. 

A visual camera is an indispensable sensor for current drones. The low cost, low power, small size of image capturing, and streaming devices make them a de facto feature for numerous drones in the market. Output of a drone's camera can be used in many ways depending of the applications. In a common scenario, the camera output is directed to the drone operator who may command the drone a new instruction based on the current visual environment via a remote controller, which serves as an agent between drone and its operator. In this work, we investigate an alternative method of controlling multirotor drones using hand gestures as the main channel of communication. We propose a framework that maps segregated images from video stream as one of five commands/gestures. The camera can capture visual instructors from the operator, which eliminates the control device, leading the way for agent-less communication. 

Haar features serve as fundamental masks to capture gradient changes in images. Each block of mask can be scaled or rotated to capture predetermined targets. These advantages allow us to detect various gestures in many sizes. Therefore, a Haar feature-based AdaBoost classifier\cite{viola2001rapid} is employed in action planner. Safety issues are also considered while the drones automatically comply with the commands initiated by operator's gestures. This project also presents a case study for image recognition-driven autonomous drones.

Our key contributions in this project include
\begin{enumerate}
\item A novel framework of drone control based on hand gestures
\item A comparison of state-of-the-art computer vision approaches in
hand gesture detection, applied on our hand gesture dataset
\item A discussion of key challenges and lessons learned from building the framework's hand gesture recognition component.
\end{enumerate}

This project uses one of many mediocre drones in the market: Parrot AR.Drone 2.0 \cite{ardrone}. Both the software library and hand gesture datasets are open-sourced at \cite{projectSite}. 


\section{Background}
Before detailing our framework, we briefly summarize related works in drone control approaches and attempts in employing gesture detection for this purpose.

\subsection{Drone Control}
Most commercial drones available on the market come with specially designed controllers, either as a dedicated signal transmitter or software applications running on users' hand-held device (such as mobile phones or tablets). In both cases, the controller sends commands with detailed movement information such as \emph{move the drone x units towards a certain direction} through wireless channels (e.g., Wi-Fi or Bluetooth). Notable products
include the DJI drones (models Phantom, Inspire, Matrice, etc.) \cite{djidrone} and Parrot's drones (models AR. Drone, Bebop, DISCO, Swing, Mambo, etc.) \cite{parrots}.

Recently there have been commercial products that introduce hand gestures
as a viable control mechanism. To capture the gestures, there are two approaches.
\begin{itemize}
\item Using specially designed gloves: The controller is mounted on a glove worn by users and detects in real time the yaw, pitch, and roll of the hand to translate into respective movements for the drone. Products include the Kd Interactive Aura Drone~\cite{Stone2017}, and the MenKind Motion Control Drone~\cite{Mughal2017}.
\item Using computer vision via the on-board camera. These devices use the on-board
camera to detect in real time where the user's hand is and respond to it in intuitive ways. Products include the DJI Spark Drone~\cite{Goldman2017}.
\end{itemize}

The first approach above presents an attempt to add new control dimensions, thus allowing more degrees of freedom to the drone controller. Instead of pressing some predefined buttons, users can move their fingers or wave their hand(s) in specific ways that are recognized by sensors installed in the glove, which are then converted into digital commands. The transmission of commands is done over radio channels, so it is the same as the traditional control paradigm. The second approach on the other hand takes a more radical leap by employing real-time image analysis, which is done on the drone itself, to recognize commands instead of sending them over radio channels.

In academia, there have been similar attempts to investigate alternative methods to control drones using body parts, such as hand gestures or full body motions. Notably, Cauchard et al.~\cite{Cauchard2015} found that when interacting with drones using body language, drone operators feel natural using gestures like those used as with a pet or other people, such as beckoning or waving. As such, natural user interfaces (NUIs) present an appealing way to enhance the user experience when interacting with drones, as compared to the traditional way of a remote-control device. In building an NUI for drone control, there are two main directions fellow researchers are working towards: with and without the help of aiding devices. 

The first involves the use of some third-party device that can recognize non-verbal gestures reliably, before mapping the detected gestures into suitable digital commands. Some such devices include the Leap Motion Controller\footnote{https://www.leapmotion.com/}~\cite{Sanders2017,Sarkar2016} and the Microsoft Kinect~\cite{Sanna2013, Pfeil2013}. 

While Leap Motion Controller is designed specifically to capture hand motions, the Kinect can capture full body motion faithfully. While this approach yields high accuracy in gesture or body motion detection, they need to be connected to a computer to work, so portability is a limiting factor.

In the second direction, body movement is detected in real-time, using machine vision, to control the drone without any additional instrument. Researchers have examined the use of eye gazes~\cite{Hansen2014}, face poses, hand gestures, and the combination of them~\cite{Nagi2014, Monajjemi2013}.

\subsection{Hand Gesture Studies}
Image-based hand gesture recognition problems have been studied extensively for decades. Twenty-four basic signs of American Sign Language are detected and classified using a boosted cascade of classifiers trained using AdaBoost and informative Haar wavelet features. In this work, Dinh et al.~\cite{dinh2006hand} have proposed a new feature called \textit{Double L} for best describing the hand gestures other than edge features, line features and edge surrounded features. Real time hand gesture detection based on bag of features and support vector machines were proposed in \cite{dardas2011real}. In training, scale invariance feature transform (SIFT) is used to extract the key-points for all training images, and vector quantization is used to map key-points from training image into bag of words after performing K-means clustering. These histograms act as feature vectors. SVM model is trained for the classification purposes. Experiments were carried out with a web camera.

A study done by Dardas et al.~\cite{dardas2011hand} detects and tracks hand gestures in cluttered backgrounds as well as in different lighting conditions. It uses skin detection and hand posture contours comparison algorithms by subtracting faces and only recognizes hand gestures using Principal Component Analysis. In each training stage, different hand gesture images with various scales, angles, and lightings are trained. The training weights are calculated by projecting training images onto the eigen vectors. During testing, the images that contained hand gestures are projected onto the eigen vectors and the testing weights are calculated. Finally, Euclidean distances are calculated between training weights and test weights to classify hand gestures.

In another work, Hu moment features used by Meng et al.~\cite{wang2013hand} proposed an algorithm for detecting the fingertip structure. First, the features which are the areas including skin region and the image, were made to differentiate the background in space of saturation, value of brightness, and hue from the skin region. Later, an algorithm to find the region of interest was implemented and fingertips were detected by approximating the contour. The seven-dimensional feature vector was created after the detection process. Finally, the distance marching criterion was used for the hand gesture recognition. This algorithm improved the accuracy by 2.7\% when compared to Hu moment feature recognition.

Detecting hand gestures in real time is a challenging task due to a few reasons, including how people perform hand gestures. Molchanov et al.~\cite{molchanov2016online}  recently addressed these challenges by a three dimensional recurrent convolutional neural network model with multi-modal input streams. The hypothesis is validated by testing multi-modal dynamic hand gesture dataset captured with depth, color and stereo infrared sensors. This system achieved an accuracy of 83.8\% in the complex dynamic hand gesture set. 

A multi-class classification approach based on Weighted Linear Discriminant Analysis and Gentle AdaBoost (GAB) algorithm was proposed by Tian et al.~\cite{tian2015hand}. In this approach, Histogram of Oriented Gradient (HoG) features are extracted arbitrarily in random locations and a multi-class cascade classifier is trained for hand gesture detection. 

\section{Proposed Framework}

\begin{figure} 
\centering
\includegraphics[width=2in]{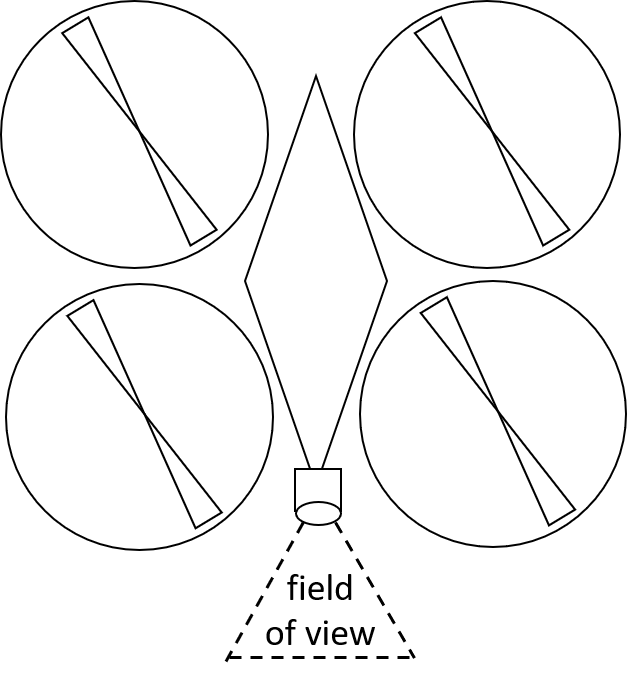}
\caption{Stylized top-down view of a quadrotor drone, facing downwards with a camera mounted at the front of the drone.}
\label{fig:quadrotor}
\end{figure}

\begin{figure*}
\centering
\includegraphics[width=5in]{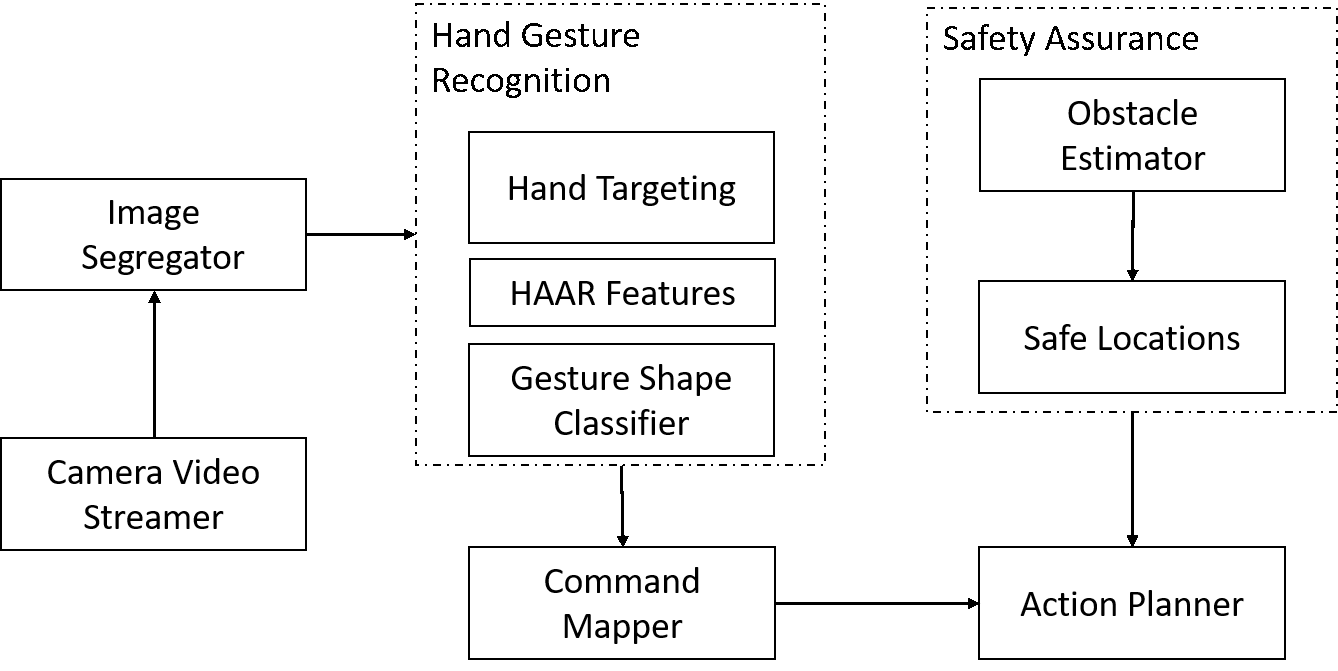}
\caption{Gesture-based drone control framework.}
\label{fig:framework}
\end{figure*}

In this section, we detail our framework of gesture-based drone control. The targeted drone types for this framework are multirotor helicopters equipped with a front-facing camera, such as the Parrot AR.Drone~\cite{ardrone}. Figure~\ref{fig:quadrotor} depicts one such drone with four rotors on the sides of the body in charge of lifting the drone off the ground and moving the drone in different directions. A camera is mounted at the front of the drone's body, which allows recording of the environment within its field of view. The framework is depicted in Figure~\ref{fig:framework}.

The video stream is constantly recorded through the on-board camera of the drone, and then segmented into sequences of still images. Each image is then analyzed through the hand gesture recognition process, which includes three main steps:  feature extraction, hand region identification, and finally gesture classification. A command mapper transforms the detected gesture into a command, such as \emph{take off}, \emph{land}, or \emph{back off}. An action planner takes the command as its input and compute the corresponding course of primitive actions to satisfy the command. While the planner is operating, it also considers the surrounding environment to avoid collision and ensure the safety for both the drone and perceived obstacles.  

\begin{figure} [!t] 
\centering
\subfloat[]{ \includegraphics[width=1in]{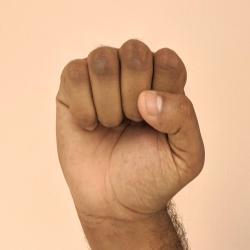} \label{l:g_fist}} 
\subfloat[]{\includegraphics[width=1in, height=1in]{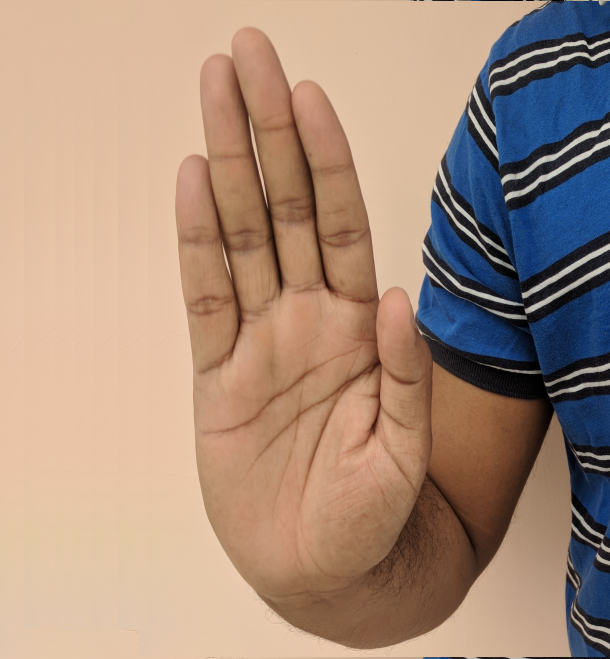} \label{l:g_palm}}
\subfloat[]{\includegraphics[width=1in]{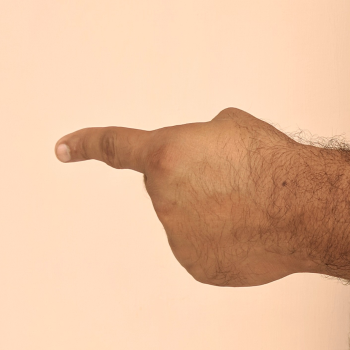} \label{l:g_go}}

\subfloat[]{\includegraphics[width=1in]{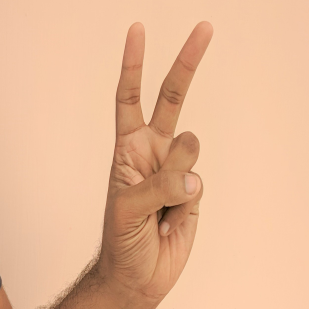} \label{l:g_v}}
\subfloat[]{\includegraphics[width=1in]{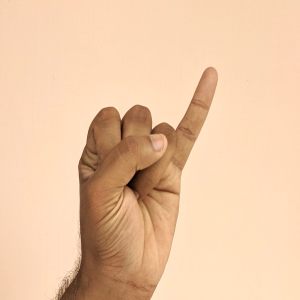} \label{l:g_l}}

\caption{The hand gestures to be classified in this study: fist with right hand (\ref{l:g_fist}), right hand palm (\ref{l:g_palm}), left pointing left hand (go symbol)(\ref{l:g_go}), left hand V-shape (\ref{l:g_v}), and left hand little finger (\ref{l:g_l}). }
\label{fig_all_ges}
\end{figure}

The hand gestures we work on are shown in Figure \ref{fig_all_ges}. Note that gestures are recognized based on certain orientation of the user's hand, i.e., either right or left hand is used for each gesture. The set of all five gestures includes fist, palm, go symbol, v-shape, and little finger. These gestures are arbitrarily picked but we made sure to have a lot of unique haar features for each carefully chosen gesture and they are very common gestures in the society and easy to pose. The reason for using only 5 gestures is to provide all basic functionalities of the drone like moving the drone right, left, backward, forward and clicking pictures. Unquestionably, more functionalities can be implemented by training more hand gestures. But the scope of this paper is focused on achieving high accuracies in mediocre drones for those basic functionalities mentioned above.

 We avoid three fingers and two fingers gestures, since they may be translated into similar Haar features, which may lead to many errors in classification step. Another example, the one finger gesture and the fingers crossed gesture may end up having similar Haar features. During the preliminary set of experiments, we decided to choose aforementioned hand gestures, assuming that they will have a new set of differentiated features for each gesture to be classified correctly. For example, the go symbol is expected to possess a separate set of Haar features when compared to the fist or the palm which in turn reduces the number of misclassified images and improves the accuracy. This does not mean Haar features end up with similar values if the gestures look alike. Therefore, we attempted to reduce one such possibility of misclassification by choosing significantly different-posing gestures. In the rest of this study, the go symbol, v-shape, and little finger gestures are indicated by GS, VS, and LF, respectively.

In order to implement the complete framework, there are a number of key challenges that we need to address, namely gesture recognition, visual variability of scene, and safety assurance of maneuver.

\subsection{Gesture Recognition}
In two related studies Viola and Jones \cite{viola2001rapid,viola2004robust} introduced Haar feature-based cascade AdaBoost classifier exclusively for frontal face recognition. Their method builds a weak classifier using extracted Haar features compiled from various sub-windows/patch of the target image. AdaBoost (Adaptive Boosting) is a weak learning algorithm and was introduced in \cite{Adaboost}. It classifies a feature vector exploiting many other subsequent learners. AdaBoost updates weights of each weak classifier at the end of each iteration in training. AdaBoost-based solutions require a set of real classifiers that learn from training dataset and map testing data to one of the classification labels. 

We used Haar features to represent each image of dataset.  Although Haar features was introduced in 1910 \cite{haar1910}, it is not popularized for image recognition problems until a broad analysis by Papageorgiou et al. \cite{haarPopular}. A Haar-based feature utilizes rectangular regions at various locations of the detection window by summing up the pixel intensities in each location of the detected window and calculates the difference between these summation values. These differences are then used to categorize the image. In our scenario, the feature extraction module uses the pattern generated by many local Haar features of a hand gesture. Later, the classifier maps feature vector of gestures either one of the existing gesture labels or as void. The reasons for choosing Haar classifiers over other algorithms are that Haar cascade has better detection rate than other feature descriptors like Hog\cite{negri2007benchmarking} in less clear images and moreover, its implementation is simple, achieves more accuracy with less training images, and consumes less memory unlike GPU-enabled image classification system like Convolutional Neural networks\cite{simonyan2014very}.

\subsection{Visual Variability of Scene} \label{sceneVariables}
\begin{figure} [!t] 
\centering
\subfloat[]{ \includegraphics[width=1in]{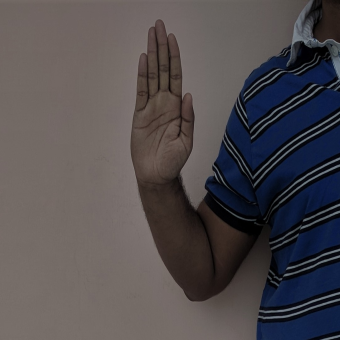} \label{l:cpalm}} 
\subfloat[]{\includegraphics[width=1in]{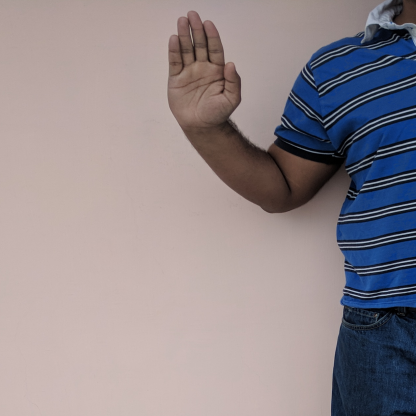} \label{l:ca-palm}}
\subfloat[]{\includegraphics[width=1in]{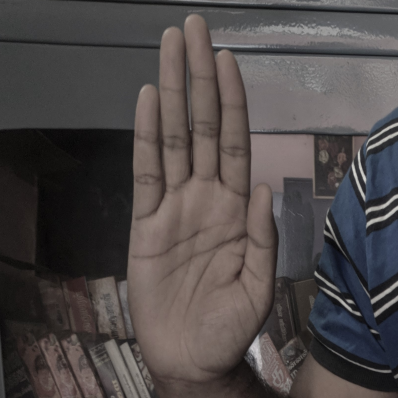} \label{l:ccd_palm}}

\subfloat[]{\includegraphics[width=1in]{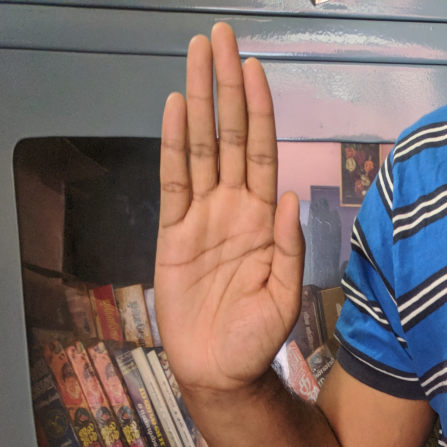} \label{l:cc_palm}}
\subfloat[]{\includegraphics[width=1in]{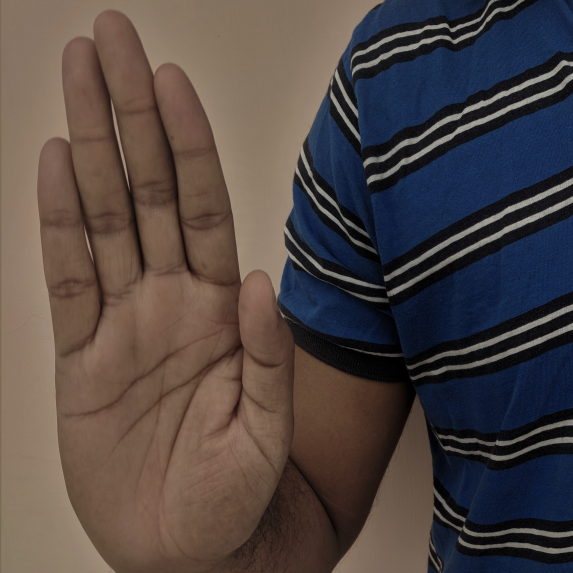} \label{l:cd-palm}}
\subfloat[]{\includegraphics[width=1in]{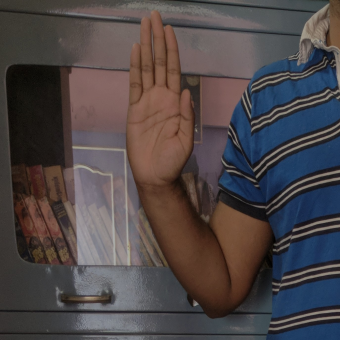} \label{l:culdim_palm}}

\caption{Demonstrating scene variability of the palm gesture. \ref{l:cpalm}: dim lit, clear background, more than 3 ft away, \ref{l:ca-palm}: well lit, clear background, more than 3 ft away, \ref{l:ccd_palm}: dim lit, cluttered background, within 3 ft, \ref{l:cc_palm}: well lit, cluttered background, within 3 ft, \ref{l:cd-palm}: dim lit, clear background, within 3 ft, \ref{l:culdim_palm}: dim lit, cluttered background, more than 3 ft away.}
\label{fig_variable_ges}
\end{figure}

The proposed study is designed for a user to control a drone in daily life, not a special laboratory environment. For this very reason, we want to empirically measure the effects of scene variability while classification framework is kept unchanged. To this end, three different visual variables are introduced to be tested: illumination, background, and distance of target gesture. The \emph{illumination} measures how well the scene is lit. In terms of illumination variable, a scene (experimental environment) is categorized in a binary way, dim lit or well lit. We did not use any special lighting tools while collecting images of the dataset. Instead, various test cases are captured under sun and/or everyday fluorescent lights. The variability of \emph{background} is expressed with one of two categories, cluttered or clear (almost blank). A user in front of a loaded bookshelf, a natural scene, or many other objects are categorized as clutter scene, whereas, a gesture posed in front of walls or doors are considered as clear background. Cluttered background problems were detailed in \cite{cluttered}. The last scene condition is basically the \emph{distance} between gesture posing hand of the user and drone's camera. The distance threshold is 3 ft. It means that while some gestures are presented within 3 ft, the others are tested more than 3 ft away. Figure \ref{fig_variable_ges} shows various scenes based on newly introduced conditions. 

\subsection{Safety}
After a gesture is recognized and converted to a command, such as \emph{move to the left}, the action planner on the drone kicks in to compute the most appropriate course of action that satisfies the recent command. In this process, it is imperative for the drone to carry out the action while ensuring safety to itself, surrounding objects, and the environment. Collision to any of these entities potentially causes serious damage to the parties involved, which is highly undesirable. In our framework, action planning module requires the drone to utilize its sensors (e.g., camera and proximity) to estimate the area where it can safely fly or hover to. 

Collision avoidance is a topic addressed in robotics. Drones are much more susceptible to external factors that cause their movements to be unstable, such as wind or air flows. Collision avoidance in drones requires additional considerations for such factors. While some approaches rely on the on-board camera for this task~\cite{Mori2013, Alvarez2016, Fu2014}, others propose the use of more advanced sensors, such as ultrasonic or laser range finders~\cite{Moses2014, Roberts2007a}. One limitation of camera-based solutions is that they may perform poorly when there are optical noises, such as in low lighting or foggy environments. Using more non-vision based sensors helps alleviate this problem, but adds more load to the overall weight of the drone, which may not always be feasible.

\section{Experiments}
Parrot AR.Drone 2.0~\cite{ardrone} is used throughout all the experiments. It is one of the early versions of the Wi-Fi controller drone, which is debuted by Parrot SA (Paris, France) in 2012. It costs around \$130 as of December 2017. AR.Drone 2.0 is equipped with 720 x 720 pixels camera, ARM Cortex A8 1 GHz 32-bit processor, Wi-Fi connectivity, gyroscope, accelerometer, magnetometer, pressure sensor, and altitude ultrasound sensor. A stylized top-down view of the AR.Drone 2.0 is shown in Figure~\ref{fig:quadrotor}.

\begin{table}
\caption{Number of images in the study dataset}
\label{l:datasize}
\centering
\begin{tabular}{ l  c c }
\multicolumn{1}{c}{\bfseries Hand Gestures } & \multicolumn{1}{c}{\bfseries Positives } & \multicolumn{1}{c}{\bfseries Negatives}  \\ [1ex]
\hline
\hline
\noalign{\smallskip}
Fist & 1570	& 900 \\ \noalign{\smallskip}
Palm & 1456 & 900 \\ \noalign{\smallskip}
GS	& 1390 & 900 \\ \noalign{\smallskip}
VS	& 1530 & 900 \\ \noalign{\smallskip}
LF 	& 1456 & 900 \\ \noalign{\smallskip}

\hline
\noalign{\smallskip}
\end{tabular}
\end{table}

Gesture recognition experiments are carried out with a 2.60 GHz CPU, 16 GB memory Ubuntu 14.04.5 LTS (Trusty Tahr) operating system. Drone control software is developed using Python 2.7 with OpenCV 3.3.0, an open-source computer vision library \cite{opencv}.

Training images are collected at resolution of 720 x 720 pixels, which are same as the drone's front camera resolution. Positive training images are hand gestures images collected from drone's front camera. Meanwhile negative training images, also called background images or background image samples, are collected randomly with the help of image search engines, which do not contain any hand gesture images. Should the size of a negative image be greater than 720 x 720 pixels, it is down-sampled to size of positive images. The number of positive and negative training samples for each gesture is given in Table \ref{l:datasize}. A total of 8302 images are used in the experiments.



We benefit from OpenCv's embedded tool to mark bounding box and location of each gesture in positive training images. It should be noted that although all five gestures are posed with same user with same right/left hand, a same gesture appears at many different orientations and scales. In a preprocessing step, their location should be marked correctly to train a classifier. OpenCV also provides an integrated annotation tool to manually describe the objects to be detected by the classifier. We created an annotation file which contains a file structure to maintain association between positive images and the coordinates of the bounding rectangles of the gestures. Following this step, we extracted features in OpenCV, which supports in creating vector representation of training images using Haar features. While generating feature vectors from the images, we specify the sample size as 20 x 20 pixels since Lienhart et al.~\cite{HaaRempirical} reported that 20 x 20 of sample size achieved the highest hit rate in a similar study. Upon extracting feature vectors, we train the boosted cascade of weak classifiers, AdaBoost, using all positive and negative feature vectors. Each of gesture classifiers are trained separately, which generates five different classification models. Once an image is streamed from the drone's camera to our software, each frame is mapped to the respective gesture or none. Training of each classifiers takes around 15 minutes because of the smaller window size (20 x 20) of Haar features extraction step. All training images and model files (in the form of .xml) are publicly available at project web site \cite{projectSite}. In the context of AdaBoost, each resulting .xml file serves as strong classifier, composed of the weighted sum of weak classifiers. The number of training stages for palm, fist, GS, VS, and LF gestures in Haar cascade classifier are reported as 4, 16, 8, 10, and 5, respectively.

\section{Discussion of Experimental Results} \label{discussion}
Individual accuracy of each gesture is detailed in Table \ref{l:accu}. This table also categorized how the classifier performs in variable scene conditions, which are described in Section \ref{sceneVariables}. The accuracy measure reported in Table \ref{l:accu} is the ratio of the correctly classified gestures to the total number of same gesture. For example, in case of the palm gesture experiments with scene variables of DL, CTB, LT-3, 4593 of 5000 palm gestures are correctly identified.

The distance is observed as the most significant scene variable. The gestures posed within 3 ft outperform significantly the gestures posed more than 3 ft away. Referring to Table \ref{l:accu}, regardless of illumination and background variability, the average accuracy of LT-3 experiments is 0.94 while that of MT-3 is 0.71. The decline of accuracy based on distance is found common amongst all gestures. One of a few sharp accuracy declines is seen in the classification of palm, where scene variable of distance is changed from LT-3 to MT-3. In this pairwise comparison, the accuracy drops from 0.97 to 0.70. The distance variable causes a comparatively mild diminishing of classification accuracy in the case of well-lit and clear background experiments, from 0.80 to 0.70.

Second and third significant scene variables are observed as background and illuminations, respectively. A cluttered background lessens accuracy in many pairwise comparisons, just as in within 3 ft, dim lit fist experiments (DL, CTB, LT-3: 0.89 while DL, CLT, LT-3: 0.91). Another example of similarly lessened accuracy is the gesture of go shape where various backgrounds of scenes are tested in well-lit and within 3 ft poses (WL, CTB, LT-3: 0.91 while WL, CLB, LT-3: 0.96).

Illumination, categorized as dim or well lit, is found the least significant scene variable. Expectedly, the effect of lighting condition is almost obvious amongst all gestures, except in a few cases of little finger and fist. The accuracy of little finger is reduced from 0.86 to 0.81 in the case of DL, CTB, LT-3 vs WL, CTB, LT-3. In the same pairwise experiments of fist, changing the illumination variable from DL to WL does not help the accuracy increase (DL, CTB, LT-3: 0.89 while WL, CTB, LT-3: 0.89). 

Overall best average classification accuracy is 0.95 and obtained with scene variables of WL, CLB, LT-3, as given at row \#7 of Table \ref{l:accu}. In summary, significance of scene variables is ordered as distance, background, and illumination, respectively.

\begin{figure} [!t] 
\centering
\subfloat[]{ \includegraphics[width=1in]{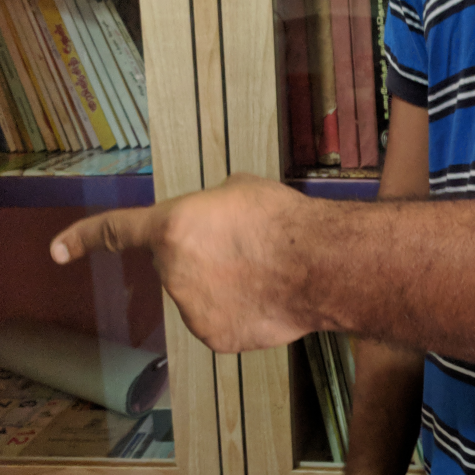} \label{l:g_as_p}} 
\subfloat[]{\includegraphics[width=1in, height=1in]{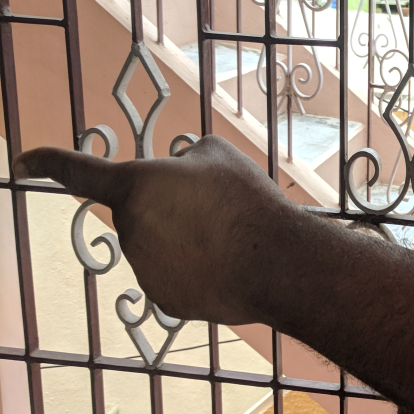} \label{l:g_as_p_2}}
\subfloat[]{\includegraphics[width=1in]{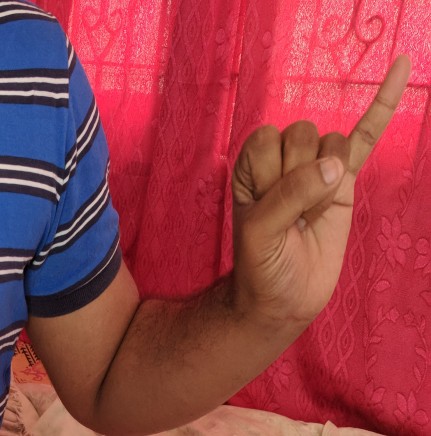} \label{l:lf_as_fist}}

\subfloat[]{\includegraphics[width=1in]{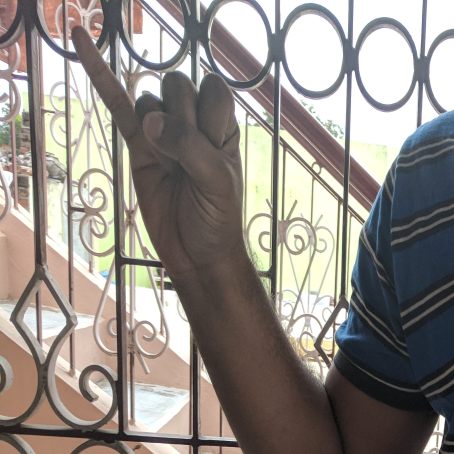} \label{l:lf_as_vs}}
\subfloat[]{\includegraphics[width=1in]{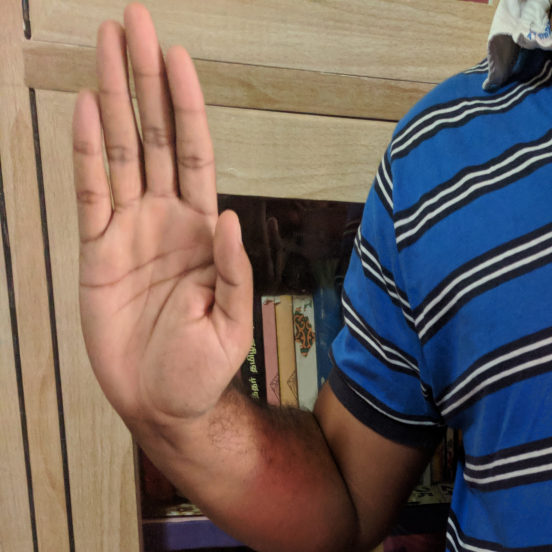} \label{l:palm_as_vs}}
\subfloat[]{\includegraphics[width=1in]{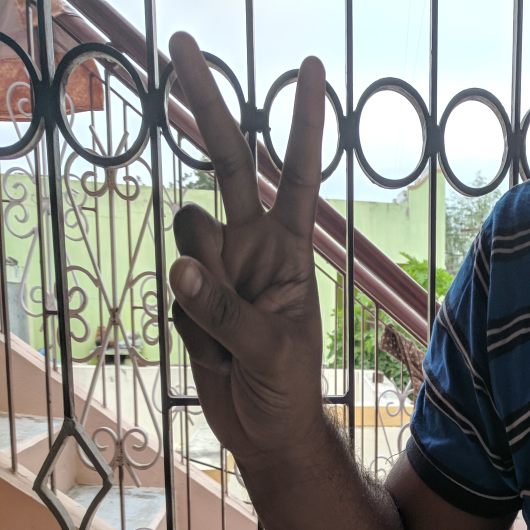} \label{l:vs_as_fist}}

\caption{Misclassified gestures. See Section \ref{discussion} for the further discussion.}
\label{fig_misclassfied}
\end{figure}

A set of misclassified gestures is depicted in Figure \ref{fig_misclassfied}. The go symbols of Figure \ref{l:g_as_p} and Figure \ref{l:g_as_p_2} are both classified as palm. The gesture of Figure \ref{l:lf_as_fist} should have been recognized as little finger but our classifier incorrectly labels it as fist. Both Figure \ref{l:lf_as_vs}  and \ref{l:palm_as_vs} are recognized as v-shape in tests. These mistakes are probably due to vertical edges in the background. As a last example of misclassification, Figure \ref{l:vs_as_fist} is a v-shape gesture; however, it is recognized as fist. 

The misclassified images give us a few insights into causes of the errors done in testing. First, the operator should be close enough to the drone for a better accuracy. This problem also involves the camera resolution of the drone and can be partially elevated with high resolution images or better cameras. We observed that Haar features are not immune to non-gesture related background patterns. This occurs because the proposed framework does not include the background removal procedure.

\begin{table}
\caption{Classification Accuracies for gesture detection}
\label{l:accu}
\centering
\begin{tabular}{ l c c c c c }
{\bfseries Test Conditions  } 
&  {\bfseries Palm } 
&  {\bfseries Fist } 
&  {\bfseries GS }  
&  {\bfseries VS } 
&  {\bfseries LF } \\ [1ex]
\hline
\hline
\noalign{\smallskip}
DL, CTB, LT-3 & 0.92 & 0.89 & 0.86 & 0.84 & 0.86 \\ 
\noalign{\smallskip}
DL, CTB, MT-3 & 0.66 & 0.70 & 0.60 & 0.65 & 0.69 \\
\noalign{\smallskip}
DL, CLB, LT-3 & 0.97 & 0.91 & 0.87 & 0.90 & 0.88 \\ 
\noalign{\smallskip}
DL, CLB, MT-3 & 0.70 & 0.74 & 0.69 & 0.65 & 0.59 \\ 
\noalign{\smallskip}
WL, CTB, LT-3 & 0.90 & 0.89	& 0.91 & 0.86 & 0.81 \\ 
\noalign{\smallskip}
WL, CTB, MT-3 & 0.69 & 0.81 & 0.73 & 0.66 & 0.70 \\
\noalign{\smallskip}
\textbf{WL, CLB, LT-3} & \textbf{0.99} & \textbf{0.99} & \textbf{0.96} & \textbf{0.95} & \textbf{0.90} \\ 
\noalign{\smallskip}
WL, CLB, MT-3 & 0.84 & 0.81 & 0.80 & 0.80 & 0.76 \\ 
\noalign{\smallskip}
\hline
\noalign{\smallskip}
\noalign{\smallskip}
\multicolumn{6}{p{205pt}}{DL: dim lit, WL: well lit, CTB: cluttered background, CLB: clear background, LT-3: within 3 ft, MT-3: more than 3 ft; GS: Go symbol, VS: V-shape, LF: little finger. The highest average accuracy settings are given in bold.}\\
\end{tabular}
\end{table}

\section{Conclusion}

Our goal of this study is to enable the hand gesture-based control mechanism with maximum possible accuracy even in mediocre drones which can be easily outperformed by the state-of-art drones due to their inbuilt high camera resolutions like 4K, 8K, 16K and 64K etc. In this empirical study, we investigated more on software development for the AR Drones. We presented an image recognition-based communication framework to control drones with hand gestures. The framework is successfully tested using a mediocre drone, Parrot AR.Drone 2.0. A set of five gestures were carefully selected to build a dataset of 8302 images. Each image is represented by a set of Haar features in cascaded AdaBoost algorithm. Classification results showed that the distance between drone and its operator is the most important indicator of success. This applies all of five gestures. Experimental tests resulted in an average accuracy of 0.90 where operated posed gestures were within 3 ft, regardless of illumination and background variability of the scene. We found that the accuracy of the framework is highest once the operator poses within 3 ft, well lit, and clear background. This controlling distance can be further improved by utilizing better cameras such as those supporting 4K or 16K resolution in the drones, which allows capturing of images with good resolution at longer distance, or implementing the same framework on state-of-art drones with better imaging capabilities. With the available HD camera in mediocre drones, the hand gesture recognition in the distance between 3 and 5 ft is highly accurate, and this controlling distance can be improved by enabling high resolution cameras in drones. With the current hand gesture-based control mechanism, we envision that drones can be sent to any feasible distances and perform operations, before flying back to the controller for further close-ranged interactions. To explore the effects of different hand poses and deviation, an in-depth statistical analysis on the applicability of the framework in different environment settings is planned for future work.

\ifCLASSOPTIONcompsoc
  \section*{Acknowledgments}
\else
  \section*{Acknowledgment}
\fi

The authors thank Texas A\&M University--Commerce Graduate School and Department of Computer Science for the travel and publication support.



%

\bibliographystyle{IEEEtran}  
\bibliography{sample}


\end{document}